\documentclass{article} 

\usepackage{aaai}
\nocopyright

\renewcommand{\land}{\wedge}
\renewcommand{\lor}{\vee}
\newcommand{\imply}{\rightarrow}

\newcommand{\biimp}{\leftrightarrow}
\newcommand{\ent}{\models}
\newcommand{\uni}{\cup}

\newcommand{\bi}{\begin{itemize}}
\newcommand{\ei}{\end{itemize}}
\newcommand{\be}{\begin{enumerate}}
\newcommand{\ee}{\end{enumerate}}
\newcommand{\beq}{\begin{equation}}
\newcommand{\eeq}{\end{equation}}
\newcommand{\bd}{\begin{description}}
\newcommand{\ed}{\end{description}}
\newcommand{\bc}{\begin{center}}
\newcommand{\ec}{\end{center}}

\newtheorem {definition}{Definition}
\newtheorem {theorem}{Theorem}
\newtheorem {property}[theorem]{Property}

\begin{document}
\bibliographystyle{aaai}

\title{Abductive and Consistency-Based Diagnosis Revisited: \newline
a Modeling Perspective}

\author{Daniele Theseider Dupr\'{e}\\
Dipartimento di Scienze e Tecnologie Avanzate\\
Universit\`{a} del Piemonte Orientale\\
Corso Borsalino 54 -- I-15100 Alessandria, Italy \\
E-mail: dtd@mfn.unipmn.it \\
}
\date{}

\maketitle

\begin{abstract}

Diagnostic reasoning has been characterized logically as consistency-based
reasoning or abductive reasoning. Previous analyses in the literature
have shown, on the one hand, that choosing the (in general more restrictive)
abductive definition may be appropriate or not, depending on the content of the
knowledge base
\cite{Console:ci:91}, and, on the other hand,
that, depending on the choice of the definition
the same knowledge should be expressed in different form \cite{Poole:94}.

Since in Model-Based Diagnosis a major
problem is finding the right way
of abstracting the behavior of the system to be modeled,
this paper discusses the relation between modeling, and in particular
abstraction in the model, and the notion of diagnosis.

\end{abstract}

\section{Introduction}

Several characterizations have been given for Model-Based Diagnosis
\cite{Console:Readings:92}.
All approaches assume that a model of the system
to be diagnosed is available: 
either a model of the correct behavior of the system,
or a model of its abnormal behavior, or both.

Diagnostic reasoning has been characterized 
as a form of nonmonotonic reasoning: either as consistency-based
reasoning, or abductive reasoning.
In the first case a set of assumptions of correct behavior must
be rejected in order to restore consistency with (abnormal) observations;
in the second case, a set of assumptions of abnormal behavior
must be introduced to entail the abnormal observations.

In \cite{Console:ci:91} the two definitions are shown to be two
extremes of a spectrum whose intermediate points may also be
relevant, depending on the assumptions about the completeness of the model.
Poole also pointed out \cite{Poole:94}
the importance of the {\em representation problem\/}
for logic-based diagnosis, i.e.\ what has to be represented about
the modeled system in order to use the different conceptualizations.

In spite of such previous work, several confusions remain in the field,
for example, the confusion of declarative issues with computational
issues, such as backward vs forward chaining along the model,
and the lack of acknowledgement that in some significant cases the
approaches are equivalent.

Moreover,
in Model-Based Diagnosis a common view is that {\em modeling} is
the problem; in particular, any model is an abstraction and
the problem is in finding the right way
of abstracting the behavior of the system to be modeled.
This is a particularly significant issue since
one of the claimed advantages of model-based systems is that they
can rely on the same model of the system for different reasoning tasks,
e.g.\ planning, diagnosis, configuration, reconfiguration after failure;
but, unfortunately, which is the right abstraction, and then
the right model, may depend on the task.

This paper, based on a general notion of prediction,
illustrates, summarizing and complementing several views in
the literature,
how the appropriate notion of diagnosis and
explanation depends on 
the predictiveness of the model.
In particular, for deterministic models abduction and consistency-based
explanation are equivalent, while
for nondeterministic models, even if at first sight consistency
seems to be the one
providing the correct diagnoses, abduction can usually be adapted to
provide the same correct diagnoses with a better (at least for someone)
notion of explanation.

\section{Basic definitions}
\label{basicdef}

In this paper we assume to rely on a {\em component-based} model
of the system to be diagnosed; such a model describes the normal
and/or abnormal behavior of the system in terms of the normal and/or
abnormal behavior of its components.
In the classical Reiter's approach \cite{Reiter:87} there was no
distinction between different abnormal behaviors, and no model or constraints
for the abnormal behavior of a component; in later papers
\cite{deKleer:89,Struss:89} the concept of {\em behavioral mode}
was introduced, and we similarly assume that:
\bi
\item
the system is composed of a set $COMPS$ of components;
\item
each component has different, mutually exclusive, behavioral modes, typically
one normal mode and several abnormal modes.
\ei
In the classical example of combinatorial circuits, a component could be
an AND gate and one of its (abnormal) behavioral modes could be
``stuck-at-0''.

In a logical representation of the model,
the fact that AND gate $a$ is in mode 
$stuck-at-0$ would be represented with the atomic formula 
$stuck-at-0(a)$, which would occur
in the formulae defining or constraining such behavioral mode,
in this case
$andgate(a) \wedge stuck-at-0(a) \imply output(a,0)$

Here we do not assume that the model is represented in logic,
it could also be, e.g., a set of qualitative equations.
In this case the equation for the ok mode of the AND gate
would be $out(a) = and(inp1(a),inp2(a)) $, with an appropriate definition
of the function $and$, while the equation corresponding to
the $stuck-at-0$ mode would be $out(a) = 0$.

In any case we are interested in the notion of {\bf parameter} in the model.
In a model including AND gates, a parameter could be, e.g., the
output of AND gate $a_1$, which in logic would be the lambda 
expression\footnote{This is the expression that, for example,
applied to 0 gives the formula $output(a_1,0)$}
$\lambda x . output(a_1,x)$ 
while in an equational model would be the variable $out(a_1)$.
A subset of the parameters is the set of {\bf observable} parameters.
Each parameter has a {\em domain} of possible values;
the parameter in the gate example has domain $\{ 0, 1 \}$ and
in general, in the qualitative models commonly used in Model-Based
Diagnosis, the domain would be finite.
The granularity of such a domain is a major modeling choice,
as, in general, is the choice of the appropriate qualitative abstraction;
a first step in providing support for this is given in \cite{Struss:qr:99}.

As mentioned above, we do not make unnecessary restrictions on
the way the model is described and, therefore, on which is the basic
inference mechanism.
E.g. the model could be a set of logical formulae, with entailment
as the inference mechanism, or a set of equations or (more generally)
constraints on finite domains, in which case constraint propagation
would be the inference mechanism.
What the model is required to provide is a notion of {\em prediction}
relating component behavior to observations as described in the following.

A diagnostic problem is characterized by a set of {\bf observations},
i.e. an assignment of values to some or all the observable parameters.
As in \cite{Console:ci:91} we distinguish between a set
$CXT$ of ``contextual'' observations (i.e.\ ``inputs'' to the system,
e.g.\ to a circuit) and a set $OBS$ of observations to be explained
by a diagnosis.

A {\bf mode assignment} is an assignment of one behavior mode to
each component in $COMPS$. 

We assume that the way the system is modeled
provides a notion of {\em prediction}, i.e.\ states whether
a mode assignment $F$ {\bf predicts} the set $S$ of values for parameter
$p$ in context $CXT$. This means that, in context $CXT$ and given
the assumptions $F$ on the behavior of components, the model of the
system implies that $p$ takes one of the values in $S$. 

For example,
in a logical framework, where the system is modeled in a set $MODEL$
of logical formulae,
for a finite set $S = \{ v_1 , \ldots , v_n \}$
of values, $F$ predicts values $S$ for $p$ in context $CXT$,
iff
$MODEL \uni F \uni CXT \ent p(v_1) \lor \ldots \lor p(v_n)$
and the same condition does not hold for any $S' \subset S$,
since we are interested in the most specific prediction.

In an equational model, given the equations $MODEL$ representing the system,
and the equations $F$ and $CXT$
corresponding to the mode assignment and the context,
the prediction for $p$ will be the set $S$ of values parameter $p$ takes
in the solutions of the system of equations $MODEL \uni F \uni CXT$.

A particularly significant case is of course
the one where $S$ is a {\em singleton}, i.e.\ the model is able to predict
an exact value for the parameter.
We will refer to this case as a {\bf deterministic prediction}.
For example, any mode assignment that gives the ``stuck-at-0''
mode to the andgate $A_1$ predicts the value $0$ for the output of $A_1$,
while of course for the ``ok'' mode the prediction will
depend on the context and on the mode of other components.

At the other extreme is the case of a fault making no prediction on
a parameter $p$, which we intend to coincide with the case where the prediction
is the whole domain of the parameter.

An {\bf observation} on a parameter $p$ will, in general, be a set of values
$O$; in a precise observation (at least as precise as the domain
granularity of $p$) such a set will be a singleton.

\begin{definition}
Given the prediction $S$ on parameter $p$ of a mode assignment $F$ in context $CXT$
and the observation $O$ for $p$, we say that:
\bi
\item
the prediction is {\bf consistent} with the observation if 
$S \cap O \neq \emptyset$
\item
the prediction {\bf implies} the observation if 
$S \subseteq O$
\ei
\end{definition}

Obviously, if prediction $S$ implies $O$ then it is consistent
with it\footnote{Since, as noticed above, we consider no prediction as
predicting the whole domain of $p$, $S$ cannot be empty.}.

Moreover, in the particular case of
a precise observation $O = \{ v \}$, a consistent prediction must include $O$,
while to imply $O$ a prediction must coincide with $O$, i.e.\ predicting value $v$
fr $p$.

In \cite{Console:ci:91} a spectrum of definitions of diagnoses
is introduced, i.e.\ a definition with a parameter $OBS^+ \subseteq OBS$ representing
the subset of the observations to be explained abductively, while for all (other)
observation consistency is required.
In the terminology introduced above, such a definition can be
reformulated as follows.

\begin{definition}
\label{defdiag}
Given diagnostic problem characterized by observations $CXT$ and $OBS$,
and a subset $OBS^+$ of $OBS$,
a {\bf diagnosis} is a mode assignment $F$ such that
\begin{enumerate}
\item
the predictions of $F$ are consistent with $OBS$
\item
the predictions of $F$ imply observations $OBS^+$
\end{enumerate}
\end{definition}

If $OBS^+ = \emptyset$ this definition is consistency-based diagnosis,
if $OBS^+ = OBS$ it is abductive diagnosis (thus imposing, in general, a stronger requirement
than consistency-based diagnosis,
reducing therefore the set of diagnoses), and all intermediate choices are possible;
in \cite{Console:ci:91} some guidelines are given for this choice, based on the
completeness of the model, i.e.\ on the fact that all the possible explanations for the
observations have been provided on the model; this corresponds to the idea of
``anticipating explanations'' in \cite{Poole:90a}, and will be referred here as
{\em backward} completeness for the reasons that will be clear in the following.

\section{Fully predictive models}

In the literature there are results on conditions that make
abductive and consistency based diagnoses coincide
\cite{Konolige:92,Poole:94}, but such results are only formulated
for logical representations where predictions are truth values
(i.e.\ a prediction on $p$ is either entailing $p$ or entailing $\neg p$).

In the context of the previous section, we introduce the notion of a model whose
prediction is deterministic on all observable parameters.

\begin{definition}
A model is {\bf fully predictive} if for any context $CXT$, for any
mode assignment $F$, for any observable parameter, $F$ makes a
deterministic prediction on $p$ in context $CXT$.
\end{definition}

Of course this condition can hardly be met if the domain of a parameter is
the set of real numbers, but it is more sensible for e.g.\ the binary domain
of combinatorial circuits and for qualitative abstractions of domains
(moreover, in some cases a model can be sensibly expressed in a form
that makes it fully predictive, as we will discuss in the next sections).

This condition corresponds to another notion of completeness of the model:
a model where the consequences of assumptions can be given non-ambiguously,
and, moreover, all the consequences of assumptions have been written down:
therefore we can refer to it as {\em forward} completeness, as opposed
to backward completeness mentioned above.

A trivial result is the following.

\begin{property}
For a fully predictive model, the choice of $OBS^+$ in definition
\ref{defdiag} is irrelevant, i.e.\ abductive and consistency-based diagnosis
coincide.

{\bf Proof.} For a fully predictive model, a prediction on $p$ is a singleton
$\{ v \}$. If it is consistent with the observation, such observation on $p$
must be a set $S$ including $v$ (or the set $\{ v \}$ itself). Therefore the
prediction implies the observation in $p$.
\end{property}

\section{Nondeterministic models}

Even with the underlying assumption that the system to be diagnosed
behaves, at a macroscopic level\footnote{I.e.\ the one of classical
physics, rather than quantum physics.},
deterministically\footnote{A faulty system can be intermittently faulty,
or, more generally, its behavior could be time-varying. Here we do not consider
the temporal dimension --- see \cite{Brusoni:aij:98} for the different ways
of taking it into account --- and we mean that the
system behaves deterministically in a time interval during which it is in the
same behavioral mode.},
a model cannot be assumed to be able to predict observations
with infinite precision.
This is the reason why we mentioned above that fully predictiveness can be
too restrictive if the domain of observations are the real numbers.

Models are, usually, a convenient abstraction of a system, and {\em qualitative} 
abstractions of real values have emerged in AI as a meaningful and (sometimes) 
useful abstraction. Such an abstraction may have pragmatical value, allowing
to represent a whole class of possible worlds at a time, and also some
cognitive value, since humans (including, sometimes, scientists and engineers) 
are able to perform qualitative inferences or even {\em tend}
to reason qualitatively,
at least in some stage in the analysis of a system.

Our view is therefore that {\em the system behaves deterministically;
it is abstraction which makes the model nondeterministic}.
Sometimes this abstraction is due to lack of knowledge, sometimes
it is just a modeling convenience.
For example, we introduce the ``flat'' fault mode for a battery
to include a whole range of values for its voltage, which would avoid
predicting exactly the voltage of a flat battery (we will return to
this example below).

The problem of 
abductive hypotheses being insufficiently predictive to entail
observations has been discussed
early in \cite{Kautz:91:named}\footnote{Dating back to 1987 
as a Technical Report.}
in the framework of plan recognition, proposing as a simple example
the fact that the plan of getting food does not explain why the agent,
whose plan we are trying to recognize, is going to a specific supermarket:
the plan will presumably only entail going to {\em some} supermarket.
For this reason Kautz rejects the explicit use of abduction
and relies on a form of closure  to achieve
the intended explanations.
Such a closure is similar to the ones in
\cite{Console:jlc:91,Konolige:92,Poole:94} (and also explanatory closure
in \cite{Reiter:91}) which in fact provide results showing the equivalence
of abduction to deduction (and consistency-based reasoning is a form
of deduction) under an appropriate closure.

A different solution can be given based on the following idea 
in \cite{Hobbs:93}.
Given an entity, e.g.\ $lube$ $oil$, more specific than another,
e.g.\ $fluid$, (so that $lube\_oil(x) \imply fluid(x)$),
if we want to explain $lube\_oil(a)$, but
our assumption only entails $fluid(a)$, 
we rely on transforming the implication
$lube\_oil(x) \imply fluid(x)$ into the equivalence
$fluid(x) \land etc(x) \biimp lube\_oil(x)$, where $etc$ is
assumable --- corresponding to the assumption that the fluid happens
to be lube oil. Assuming $etc(a)$ allows entailing $fluid(a)$.

Similar ideas are part of the representation methodology
in \cite{Poole:90a}, in particular the idea of
``anticipating explanations'' which is also
combined with the idea of `` parametrizing assumptions''.
The latter is used for example to represent the model of a flat battery
with
\[ battery(B) \land flat(B,V) \imply voltage(B,T) \]
\[ flat(B,V) \imply 0 \leq V \leq 1.2 \]
where $flat$ is assumable, so that e.g.\ $voltage(b1,0.8)$, given $battery(b1)$,
is explained by $flat(b1,0.8)$ {\em which is also consistent} with the
constraint.
This achieves the result of being able to describe concisely a class
of faults as well as reasoning, when necessary, on a specific instance
of the class.

The parametrized assumption
methodology has been adopted e.g.\ in \cite{Ng:92} for the same
reason, and in \cite{Brusoni:aij:98} for {\em temporal} constraints
between the temporal extent of ``causes'' and ``effects'',
or, in general, explanans and explanandum. In that case, in fact,
the same problem arises unless the temporal extent of the cause
uniquely determines the one of the effect.

In general, this problem arises with any case where 
there are non-functional constraints
between parameters of the explanans and those of the explanandum,
i.e.\ the constraints do not allow to entail exact values for
the latter even given exact values for the former.

In \cite{Cordier:98}, however, the parametrizing assumptions methodology
is rejected due to the potential proliferation of hypotheses,
and a notion of explanation is introduced where ``explaining'' is
not entailing, in particular, having a prediction which is
an abstraction of the observation is sufficient.
Further, a notion of {\em conditional explanation} is introduced where
prediction and observation must have something that is more specific
than both (e.g.\ $A$ for $A \lor B$ and $A \lor C$).
In terms of predicting and observing values for parameters (as in the
definitions above) this coincides with prediction and observation
having an intersection, i.e.\ with consistency-based explanation.

\section{Qualitative deviations}

In this section we discuss abductive and consistency-based explanations
in the context of a representational abstraction 
we have used in recent years for modeling systems in the 
Vehicle Model-Based Diagnosis (VMBD) project:
qualitative deviations \cite{Struss:dx96,Cascio:aicomm:99,Theseider:DX98}.

The system is modeled in terms of differential equations 
that include appropriate
parameters for components, 
whose values correspond to different (correct or faulty)
behavior modes of the component. 
From these equations\footnote{Due to the qualitative abstraction,
the exact form of the quantitative equation is not necessary
to build  the qualitative model; it could however be necessary
for fault detection.}, corresponding equations for
qualitative deviations are derived:

\begin{itemize}
\item
for each variable x, 
its deviation $\Delta x(t)$ is defined as $\Delta x(t) = x(t) - x_{ref}(t)$,
where $x_{ref}(t)$ is a reference behavior (one choice is to consider
the correct behavior of all the system as the reference behavior);
\item
from any equation $A=B$, 
the corresponding equation $\Delta A$ $=$ $\Delta B$ is derived;
\item
finally, the corresponding {\em sign} equation
[$\Delta A$] = [$\Delta B$]
is derived; it equates the {\em signs} of the two deviations.
There are rules for expressing this equation in terms of
signs of deviations of individual variables rather than expressions.
\end{itemize}
   
These models are useful to express concisely a number of dependencies.
For example, for a tank containing a liquid, with an input flow
$in$ and an output flow $out$, the equation
\[ \partial \Delta level = [ \Delta in ] \ominus [ \Delta out ] \]
where $\partial \Delta level$ is the sign of
the derivative of the deviation of $level$
and $\ominus$ is subtraction in the sign algebra,
expresses how the level of liquid deviates from the expected value
in a wide (and exhaustive) range of cases.
In particular, the reference behavior of the system need not be a stable
state: even in such a behavior, flows and level may be continuously varying.
What the model states is, for example, that: 
\bi
\item
starting from the reference behavior,
the level does not
deviate from it, if the flows do not;
\item
if the outflow has no deviation and the inflow has a negative deviation,
the level will start deviating negatively ---
this includes the cases where it decreases instead of being constant,
decreases more than expected, increases less than expected,
becomes constant instead of increasing, and decreases instead of increasing.
\ei
But what if the outflow also has a negative deviation? Subtracting
a negative number from another can give a positive, negative or zero result.
Again, the qualitative abstraction makes the model nondeterministic,
at least in some case.

We do not discuss here an appropriate notion of diagnosis for dynamic
systems (see \cite{Theseider:DX98}), nor a complete model for
a system including the tank, but suppose that the inflow is due to
a pump (so that a pump fault makes the inflow negatively deviated)
and the outflow is regulated by a control system.

If the pump fault occurs,
there is a negative deviation of the inflow, so a negative deviation
of the level's derivative, and then of the level; 
then (we omit the model) the control system reacts
with a negative deviation of the outflow, and, given the qualitative
model, any result is possible for the level deviation's derivative.
Therefore the fault will not entail any observation on the further
trend of the level; this means that it would not be an abductive
diagnosis for any observation, while it 
would be a consistency-based
diagnosis for any observation. 
But would it be considered a good explanation? 

We can do something better without abandoning the convenience of
qualitative abstractions and using abduction.
Suppose e.g.\ we observe that the level deviation's derivative changes
sign: the pump fault is consistent, but does not predict this;
what is the least presumptive assumption that predicts this?
The assumption that $\Delta out < \Delta in $ (which for absolute
values means $|\Delta out| > |\Delta in| $), so that 
$[ \Delta in ] \ominus [ \Delta out ]$ gives '+'.
A natural language expression of this explanation is that the
control system is compensating for the fault.
Notice that the assumption
$\Delta out < \Delta in $ is still a qualitative assumption
and could correspond to a natural way of interpreting what is going on
in the system.

Thus, in general,
for all cases where the result of sign operators
is ambiguous, we can introduce assumptions
that make the result unambiguous, e.g.\ for the sum $[a] \oplus [b]$
where $a$ and $b$ have opposite signs, we can use the assumptions
$|a|<|b|$, $|a|=|b|$, $|a|>|b|$ to predict values '-',0,'+'
respectively.

\section{Conclusions}

In this paper we have reviewed several points of view in the literature
on the relation of the notion of diagnosis with properties of
the model of the system to be diagnosed, in particular, properties
related to the abstractions in the model with respect to the real
system.

We have pointed out that ideas that have appeared in the literature
on this subject can be successfully integrated with modeling
abstractions that have become widely used more recently.

The problem of identifying the right modeling abstraction for a task
is far from being solved, but some work on this is being done
in the model-based reasoning community (see e.g. \cite{Brusoni:aij:98}
for the temporal dimension in diagnosis) and we expect in the near future
more work in this direction, starting e.g.\ from the results
in \cite{Struss:qr:99}.

\end{document}